\documentclass{IEEEcsmag}

\usepackage[colorlinks,urlcolor=blue,linkcolor=blue,citecolor=blue]{hyperref}
\expandafter\def\expandafter\UrlBreaks\expandafter{\UrlBreaks\do\/\do\*\do\-\do\~\do\'\do\"\do\-}
\usepackage{upmath,color}

\jvol{XX}
\jnum{XX}
\paper{8}
\jmonth{December}
\jname{Computing in Science \& Engineering}
\jtitle{An Intuitive Tutorial to Gaussian Process Regression}
\pubyear{2023}

\usepackage{amsmath,amsfonts}
\usepackage{algorithmic}
\usepackage{array}
\usepackage{textcomp}
\usepackage{stfloats}
\usepackage{url}
\usepackage{verbatim}
\usepackage{graphicx}

\usepackage[colorlinks,urlcolor=blue,linkcolor=blue,citecolor=blue]{hyperref}
\usepackage{color}
\usepackage{amssymb}
\usepackage{amscd}
\usepackage{float}
\usepackage{subfig}
\usepackage{nccmath}
\usepackage{siunitx}
\usepackage{booktabs}
\usepackage{url}

\begin{document}

\sptitle{Feature Article: Introduction to Gaussian Processes}

\title{An Intuitive Tutorial to Gaussian Process Regression}

\author{Jie Wang}
\affil{University of Waterloo, Waterloo, ON, N2L~3G1, Canada}



\markboth{THEME/FEATURE/DEPARTMENT}{THEME/FEATURE/DEPARTMENT}

\begin{abstract}This tutorial aims to provide an intuitive introduction to Gaussian process regression (GPR). GPR models have been widely used in machine learning applications due to their representation flexibility and inherent capability to quantify uncertainty over predictions. The tutorial starts with explaining the basic concepts that a Gaussian process is built on, including multivariate normal distribution, kernels, non-parametric models, and joint and conditional probability. It then provides a concise description of GPR and an implementation of a standard GPR algorithm. In addition, the tutorial reviews packages for implementing state-of-the-art Gaussian process algorithms. This tutorial is accessible to a broad audience, including those new to machine learning, ensuring a clear understanding of GPR fundamentals.
\end{abstract}

\maketitle

\chapteri{G}aussian Process is a key model in probabilistic supervised machine learning, widely applied in regression and classification tasks. It makes predictions incorporating prior knowledge (kernels) and provides uncertainty measures over its predictions \cite{Rasmussen2006}. Despite its broad application, understanding GPR can be challenging, especially for professionals outside computer science, due to its reliance on complex concepts like multivariate normal distribution, kernels, and non-parametric models.

This tutorial aims to explain GPR in a clear, accessible way, starting from fundamental mathematical concepts including multivariate normal
distribution, kernels, non-parametric models, and joint and conditional probability. To facilitate an intuitive understanding, the tutorial extensively utilizes plots and provides practical code examples, available at \url{https://github.com/jwangjie/Gaussian-Process-Regression-Tutorial}. This tutorial is designed to make GPR accessible to a diverse audience, ensuring that even those new to the field can grasp its core principles.

\section{MATHEMATICAL BASICS}

\begin{figure}
    \centering
    \subfloat[Data point observations]{{\includegraphics[trim=2.0cm 0.8cm 3.2cm 1.8cm, width=6.3cm]{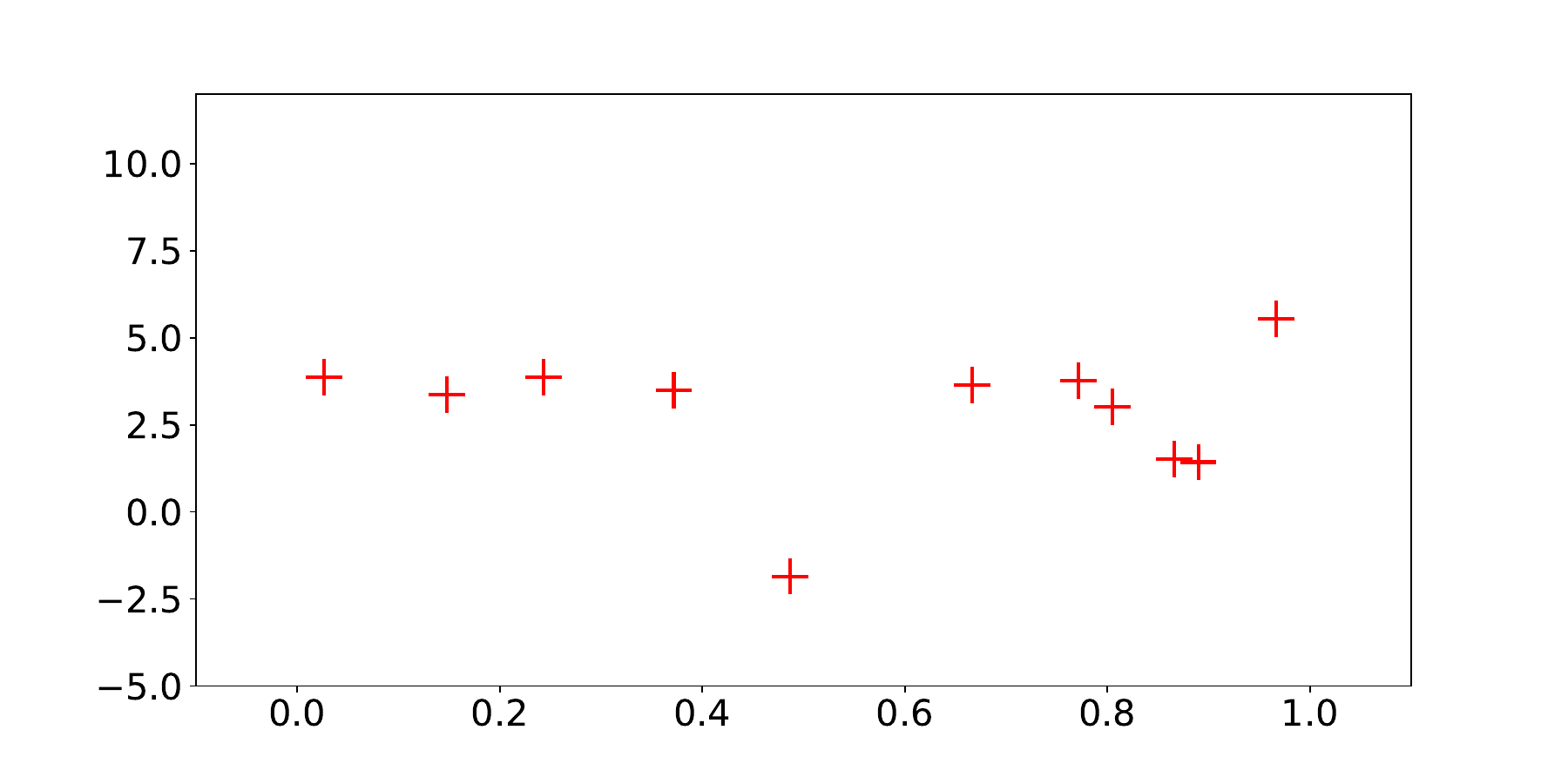} }}
    \qquad
    \subfloat[Five possible functions by GPR]{{\includegraphics[trim=2.0cm 0.8cm 3.2cm 1.8cm, width=6.3cm]{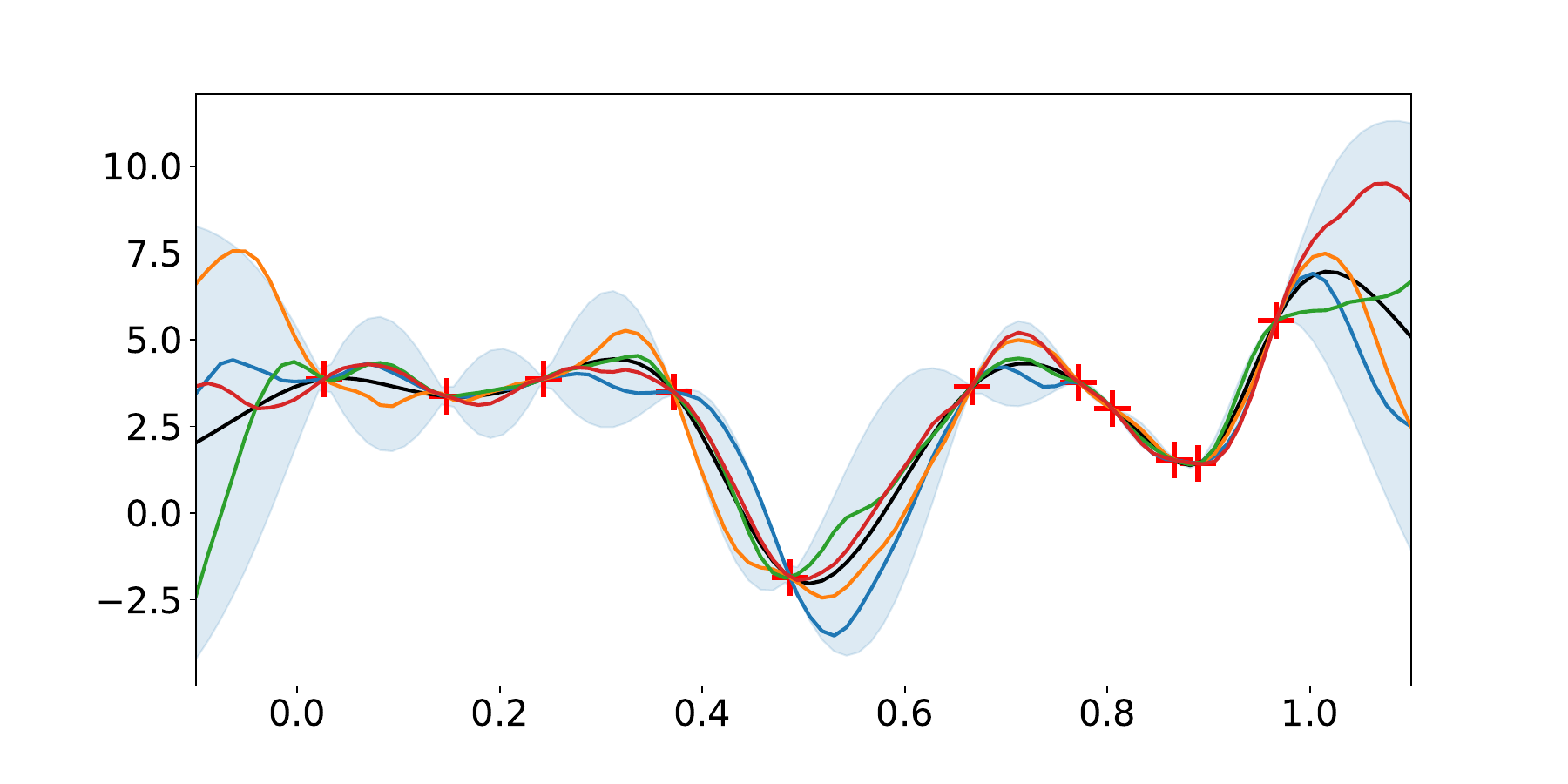} }}
    \caption{A regression example: (a) Observed data points, (b) Five sample functions fitting the observed data points.}%
    \label{FIG:1}
\end{figure}

This section explains the foundational concepts essential for understanding Gaussian process regression (GPR). We start with the Gaussian (normal) distribution, followed by an explanation of multivariate normal distribution (MVN) theories, kernels, non-parametric models, and the principles of joint and conditional probability. The objective of regression is to formulate a function that accurately represents observed data points and then utilize this function for predicting new data points. Considering a set of observed data points depicted in Fig. \ref{FIG:1}(a), an infinite array of potential functions can be fitted to these data points. Fig. \ref{FIG:1}(b) illustrates five such sample functions. In GPR, Gaussian processes perform regression by defining a distribution over this infinite number of functions \cite{ghahramani2011tutorial}.
\subsection{Gaussian Distribution}

\begin{figure}
	\centering
		\includegraphics[trim=0.4cm 0.1cm 1.4cm 0.6cm, width=0.72\columnwidth]{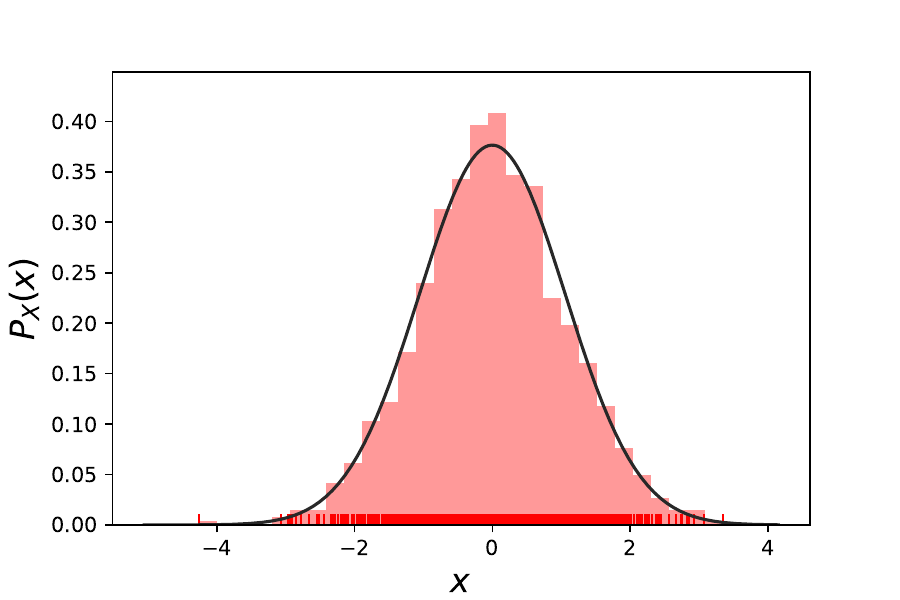}
    	\caption{Visualization of 1000 normally distributed data points as red vertical bars on the $X$-axis, alongside their PDF plotted as a two-dimensional bell curve.}
    	\label{FIG:2}
\end{figure}

A random variable $X$ is Gaussian or normally distributed with mean $\mu$ and variance $\sigma^2$ if its probability density function (PDF) is \cite{Murphy2012}:
\begin{ceqn}
    \begin{align}
       P_X(x) = \frac{1}{\sqrt{2 \pi} \sigma} exp{\left(-\frac{{\left(x - \mu \right)}^{2}}{2 \sigma^{2}}\right)} \ . \nonumber
    \end{align}
\end{ceqn}
Here, $X$ represents random variables and $x$ is the real argument. This normal distribution of $X$ is usually represented by $ P_X(x) ~ \sim\mathcal{N}(\mu, \sigma^2)$. The PDF of a uni-variate normal (or Gaussian) distribution was plotted in Fig. \ref{FIG:2}, where 1000 points from a uni-variate normal distribution were randomly generated and plotted along the $X$ axis.

\begin{figure}
	\centering
	\includegraphics[width=0.85\columnwidth]{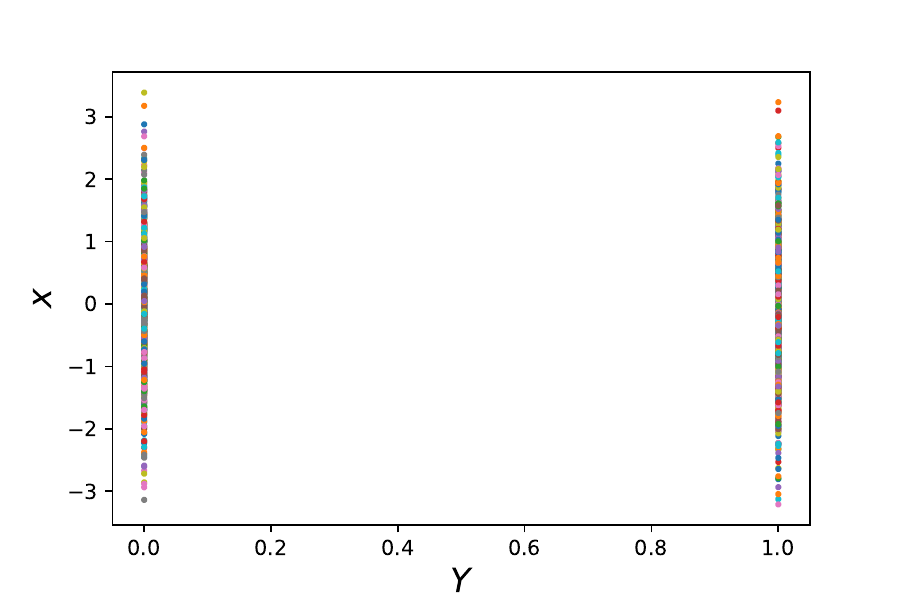}
	\caption{Two independent uni-variate Gaussian vector points plotted vertically within the $Y, x$ coordinates space.}
	\label{FIG:3}
\end{figure}

These randomly generated data points can be expressed as a vector $x_1=[x_1^1, x_1^2, \ldots, x_1^n]$. By plotting the vector $x_1$ on a new $Y$ axis at $Y = 0$, we projected the points $[x_1^1, x_1^2, \ldots, x_1^n]$ into a different space shown in Fig. \ref{FIG:3}. We did nothing but vertically plot points of the vector $x_1$ in a new $Y, x$ coordinates space. Similarly, another independent Gaussian vector $x_2=[x_2^1, x_2^2, \ldots, x_2^n]$ can be plotted at $Y = 1$ within the same coordinate framework, as demonstrated in Fig. \ref{FIG:3}. It's crucial to remember that both $x_1$ and $x_2$ are a uni-variate normal distribution depicted in Fig. \ref{FIG:2}. 

\begin{figure}
    \centering
    \subfloat[Two Gaussian vectors]{{\includegraphics[trim=0.3cm 0.1cm 1.4cm 0.7cm, width=0.72\columnwidth]{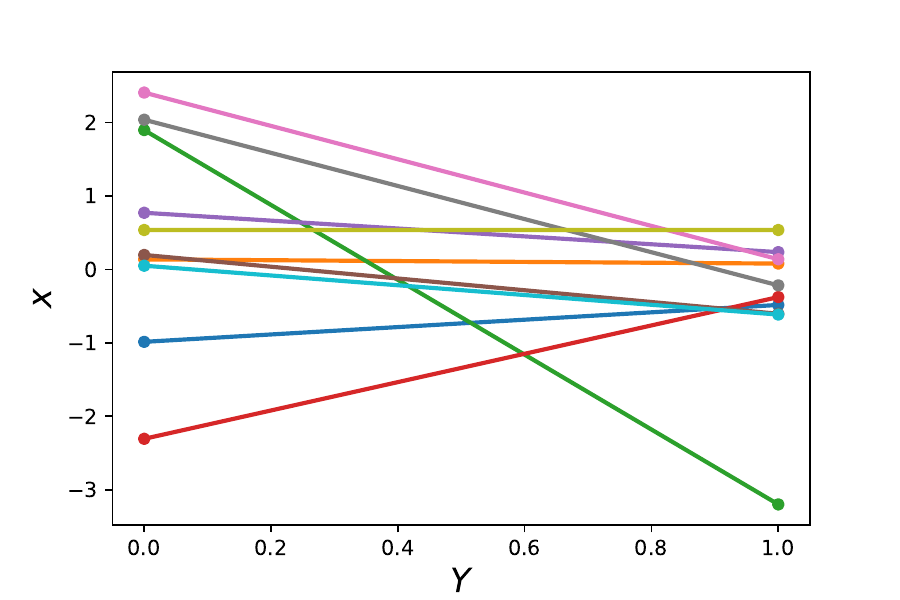} }}
    \qquad 
    \subfloat[Twenty Gaussian vectors]{{\includegraphics[trim=0.45cm 0.1cm 1.6cm 0.7cm, width=0.7\columnwidth]{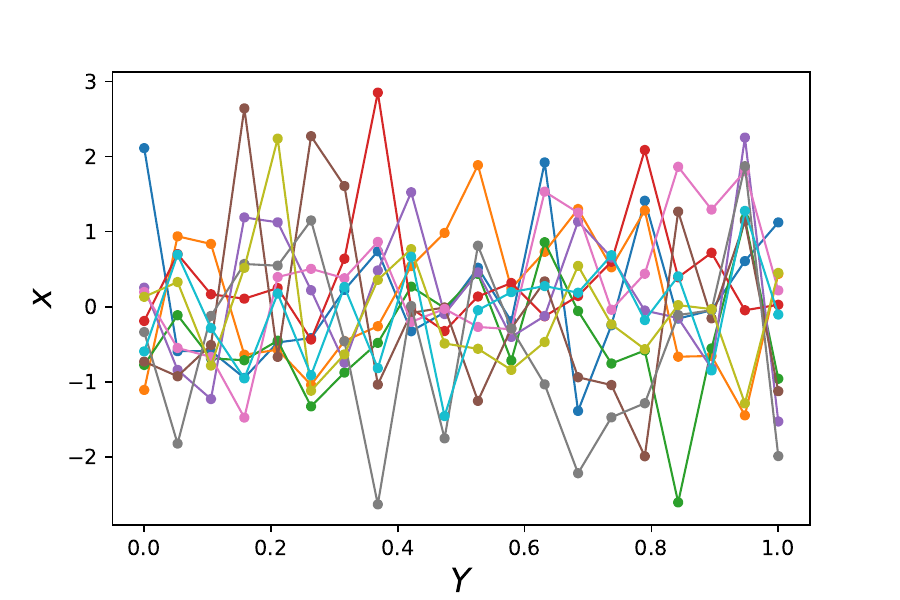} }}
    \caption{Connecting points of independent Gaussian vectors by lines: (a) Ten randomly selected points in vectors $x_1$ and $x_2$, (b) Ten randomly selected points in twenty vectors $x_1, x_2, \ldots, x_{20}$ .}
    \label{FIG:4}
\end{figure}

Next, we selected 10 points randomly in vector $x_1$ and $x_2$ respectively and connected these points in order with lines as shown in Fig. \ref{FIG:4}(a). These connected lines look like linear functions spanning within the $[0, 1]$ domain. We can use these functions to make predictions for regression tasks if the new data points are on (or proximate to) these linear lines. However, the assumption that new data points will consistently lie on these linear functions often does not hold. If we plot more random generated uni-variate Gaussian vectors, say 20 vectors like $x_1, x_2, \ldots, x_{20}$ within $[0, 1]$ interval, and connecting 10 randomly selected sample points of each vector as lines, we get 10 lines that look more like functions within $[0, 1]$ shown in Fig. \ref{FIG:4}(b). Yet, we still cannot use these lines to make predictions for regression tasks because they are too noisy. These functions must be smoother, meaning input points that are close to each other should have similar output values. These ``functions'' generated by connecting points from independent Gaussian vectors lack the required smoothness for regression tasks. Therefore, it is necessary to correlate these independent Gaussians, forming a joint Gaussian distribution, as described by the theory of multivariate normal distribution.  

\subsection{Multivariate Normal Distribution}

It is quite usual and often necessary for a system to be described by more than one feature variable $(x_1, x_2, \ldots, x_D)$ that are correlated to each other. If we would like to model these variables all together as one Gaussian model, we need to use a multivariate Gaussian/normal (MVN) distribution model \cite{Murphy2012}. The PDF of a D-dimensional MVN is defined as \cite{Murphy2012}:
\begin{equation}
    \mathcal{N}(x | \mu,\Sigma) = \dfrac{1}{(2\pi)^{D/2}|\Sigma|^{1/2}}\exp\left[-\dfrac{1}{2}(x-\mu)^\mathsf{T} \Sigma^{-1}(x-\mu)\right] , \nonumber
\end{equation}
Here, $D$ represents the number of the dimensionality, $x$ denotes the variable, $\mu=\mathbb{E}[x] \in \mathbb{R}^D$ is the mean vector, and $\Sigma=\text{cov}[x]$ is the $D \times D$ covariance matrix. The $\Sigma$ is a symmetric matrix that stores the pairwise covariance of all jointly modeled random variables, with $\Sigma_{ij}=\text{cov}(y_i,y_j)$ as its $(i,j)$ element. 

\begin{figure}
    \centering
    \subfloat[3-D bell curve]{{\includegraphics[trim=1.6cm 1.0cm 2.6cm 1.0cm, width=0.8\columnwidth]{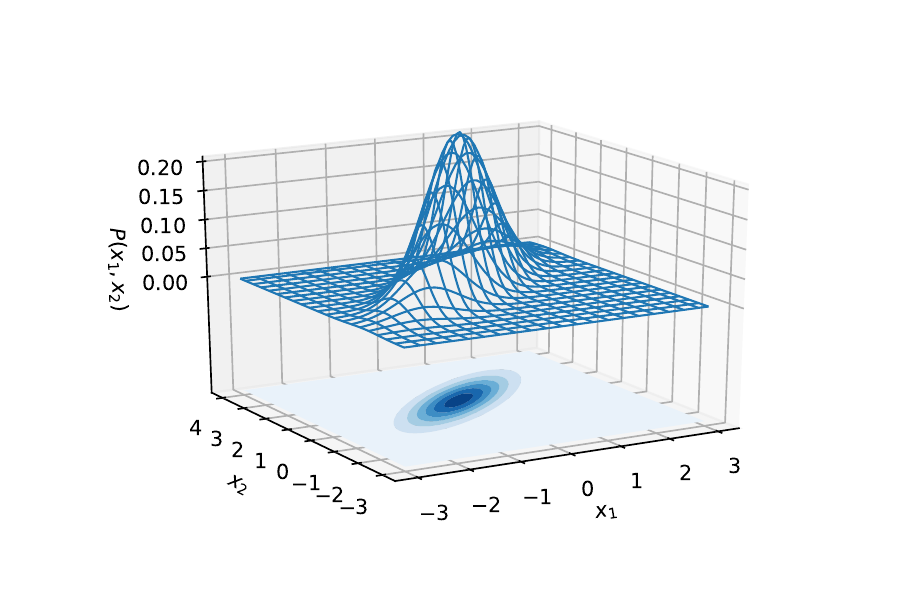} }}
    \qquad
    \subfloat[2-D ellipse contours]{{\includegraphics[trim=0.3cm 0.0cm 0.8cm 0.2cm, width=0.6\columnwidth]{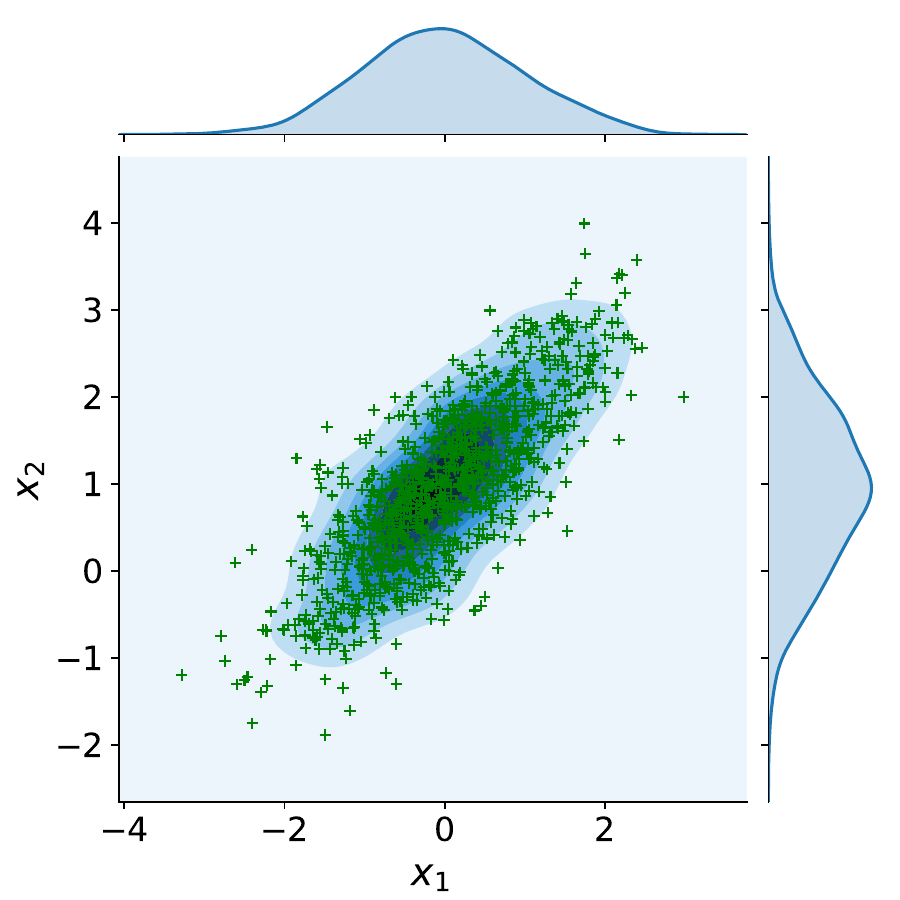} }}
    \caption{BVN PDF visualization: (a) a 3-D bell curve with the height representing the probability density, (b) 2-D ellipse contour projections showing the correlation between $x_1$ and $x_2$ points.}%
    \label{FIG:5}
\end{figure}

A bi-variate normal (BVN) distribution offers a simpler example to understand the MVN concept. A BVN distribution can be visualized as a three-dimensional (3-D) bell curve, where the vertical axis (height) represents the probability density, as shown in Fig. \ref{FIG:5}(a). The ellipse contours on the $x_1, x_2$ plane, illustrated in Fig. \ref{FIG:5}(a) and \ref{FIG:5}(b), are the projections of this 3-D curve. The shape of ellipses shows the correlation degree between $x_1$ and $x_2$ points, i.e. how one variable of $x_1$ relates to another variable of $x_2$. The function $P(x_1, x_2)$ denotes the joint probability density of $x_1$ and $x_2$. 
For a BVN, the mean vector $\mu$ is a two-dimensional vector $\begin{bmatrix} \mu_1 \\ \mu_2 \end{bmatrix}$, where $\mu_1$ and $\mu_2$ represent the independent means of $x_1$ and $x_2$, respectively. The covariance matrix is $\begin{bmatrix} \sigma_{11} & \sigma_{12} \\ \sigma_{21} & \sigma_{22} \end{bmatrix}$, with the diagonal terms $\sigma_{11}$ and $\sigma_{22}$ being the independent variance of $x_1$ and $x_2$, respectively. The off-diagonal terms, $\sigma_{12}$ and $\sigma_{21}$ represent correlations between $x_1$ and $x_2$. The BVN is expressed as: 
\begin{ceqn}
    \begin{align}
       \begin{bmatrix} x_1 \\ x_2 \end{bmatrix} \sim \mathcal{N}\left(\begin{bmatrix} \mu_1 \\ \mu_2 \end{bmatrix}, \begin{bmatrix} \sigma_{11} & \sigma_{12}  \\ \sigma_{21} & \sigma_{22} \end{bmatrix}\right) = \mathcal{N}(\mu, \Sigma) \ . \nonumber
    \end{align}
\end{ceqn}

\begin{figure}
	\centering
	\includegraphics[trim=0.4cm 0.4cm 0.6cm 0.8cm, width=0.8\columnwidth]{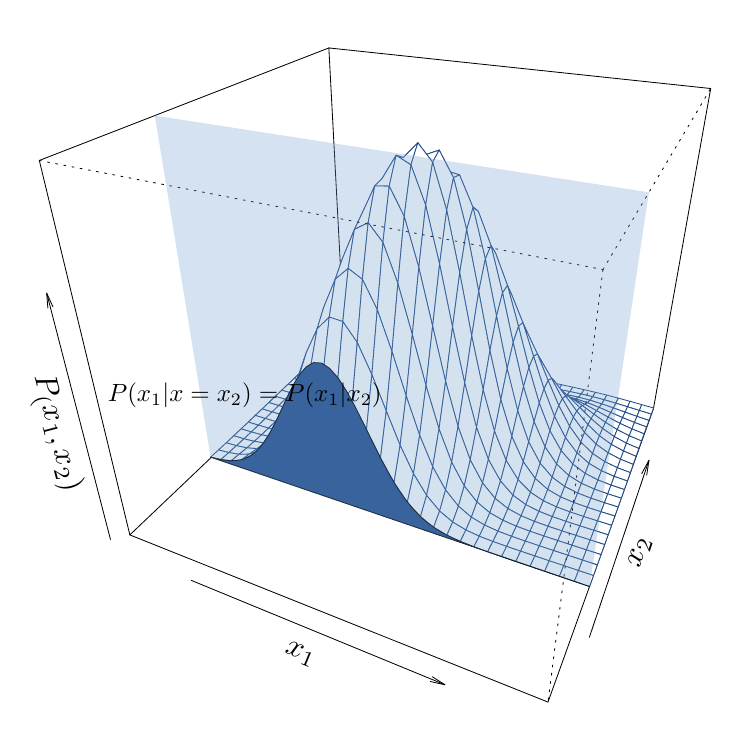}
	\caption{The conditional probability distribution $P(x_1 \vert \, x_2)$ obtained by cutting a slice on the PDF 3-D bell curve of a BVN.}
	\label{FIG:6}
\end{figure}

It is intuitively understandable that we need conditional probability rather than joint probability for regression tasks. If slicing the 3-D bell curve of a BVN at a certain constant point, as shown in Fig. \ref{FIG:5}(a), we can obtain the conditional probability distribution $P(x_1 \vert \, x_2)$, with $x=x_2=\text{constant}$, shown in Fig. \ref{FIG:6}. This conditional distribution is also Gaussian \cite{Rasmussen2006}.

\subsection{Kernels}

Having introduced the MVN distribution, we want to smooth the functions in Fig. \ref{FIG:4}(b) for regression tasks. The kernel, or covariance function, plays a pivotal role in this smoothing process, encapsulating our prior knowledge about the functions we aim to model. In regression, we desire the predictions to be smooth and logical: similar inputs should yield similar outputs. For example, consider two houses, A and B, with comparable size, location, and features; we expect their market prices to be similar. A natural measure of `similarity' between two inputs is the dot product $A \cdot B = \lVert A \rVert \lVert B \rVert \text{cos}\theta$, where $\theta$ is the angle between two input vectors. Smaller angles, indicating high similarity, correspond to larger dot products, and vice versa.

Imagine a scenario in which we could lift our house into a `magical' space, where doing this dot product becomes more powerful and tells us even more about how similar our houses are. This magical space is called ``feature space''. The function that helps us do this lift and enhanced compression in the feature space is named as ``kernel function'', denoted as $k(x,\ x^\prime)$. We do not actually move our data into this new high-dimensional ``feature space''  (that could be computationally expensive); instead, the kernel function facilitates the comparison of data though providing us the same dot product result as if we had done so. This is known as the famous ``kernel trick''. Formally, the kernel function $k(x,\ x^\prime)$ computes the similarity between data points in a high-dimensional feature space without explicitly transforming the inputs \cite{Rasmussen2006}. Instead of directly computing the dot product of transformed inputs, $\langle \phi(x), \phi(x') \rangle$, with $\phi$ being the feature mapping function, the kernel function accomplishes the same result in a computationally efficient manner. 

\begin{figure}
    \centering
    \subfloat[Ten samples of the 20-VN prior with an identity kernel]{{\includegraphics[trim=0.9cm 0.4cm 1.6cm 0.6cm, width=0.7\columnwidth]{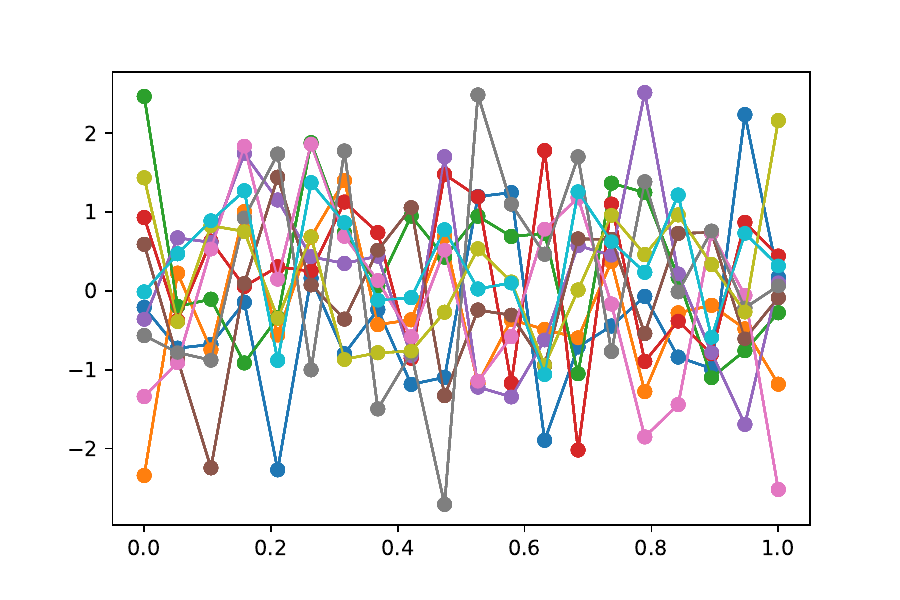} }}
    \qquad
    \subfloat[Ten samples of the 20-VN prior with a RBF kernel]{{\includegraphics[trim=0.9cm 0.4cm 1.6cm 0.6cm, width=0.7\columnwidth]{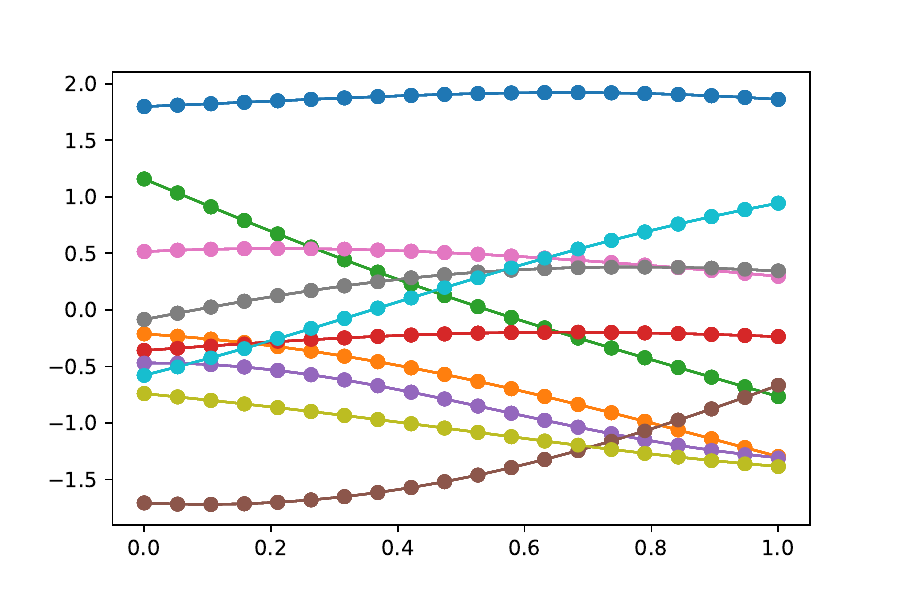} }}
    \caption{Samples of twenty-variate normal (20-VN) distribution kernelized prior functions: (a) Ten 20-VN with identity covariance, (b) Ten 20-VN with RBF covariance.}
    \label{FIG:7}
\end{figure}

The squared exponential (SE) kernel, also known as the Gaussian or Radial Basis Function (RBF) kernel, is widely used in Gaussian processes due to its exceptional properties \cite{duvenaud2014automatic}.  It is recognized for its adaptability across various functions. Additionally, every function in its prior is smooth and infinitely differentiable, leading to naturally smooth and differentiable model predictions. The SE kernel function is defined as \footnote{This is a simplified SE kernel without hyperparameters for simplicity. The general SE kernel will be further explained in Sec. Hyperparameters Optimization.}:
\begin{ceqn}
    \begin{align}
       \text{cov}(x_i, x_j)=\exp\left(-~\frac{(x_i-x_j)^2}{2}\right) \ . \nonumber
    \end{align}
\end{ceqn}
In Fig. \ref{FIG:4}(b), we plotted 20 independent Gaussian vectors by connecting 10 randomly selected sample points from each vector in order by lines. Instead of plotting 20 independent Gaussian, we can generate 10 twenty-variate normal (20-VN) distributions with an identity covariance function as shown in \ref{FIG:7}(a). It is the same as Fig. \ref{FIG:4}(b) due to the absence of correlations among points by using identity as its kernel function. Employing an RBF kernel as the covariance function, on the other hand, we got smooth lines observed in \ref{FIG:7}(b).

\begin{figure}
	\centering
		\includegraphics[width=\columnwidth]{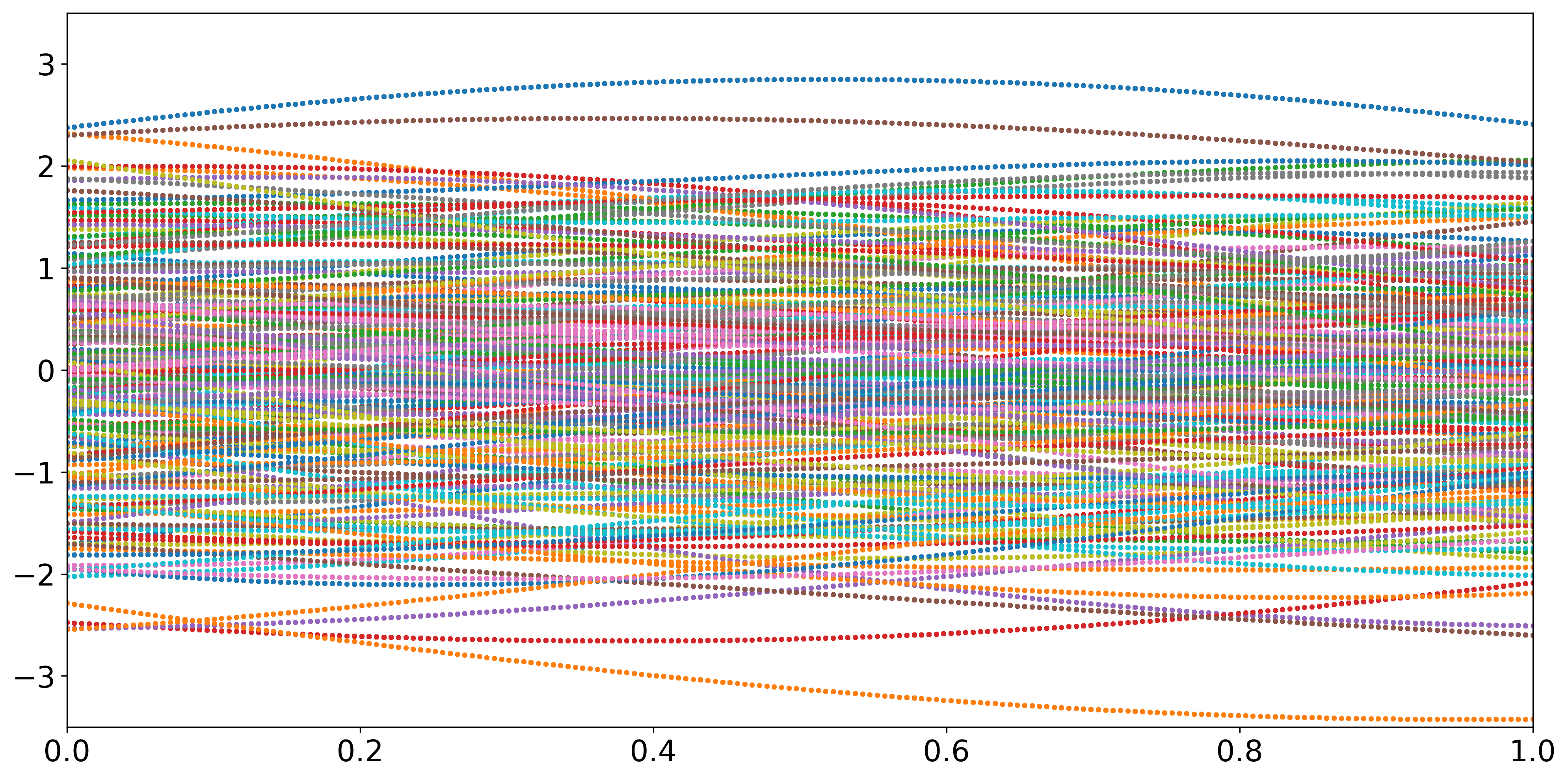}
	\caption{Two hundred kernelized prior functions from a two hundred-variate normal distribution.}
	\label{FIG:8}
\end{figure}

By integrating covariance functions, we obtain smoother lines, and they start to look like functions. It is natural to consider continuing to increase the dimension of MVN. Here, dimension refers to the number of variables in the MVN. When the dimension of MVN becomes larger, the region of interest will be filled up with more points. When the dimension reaches infinity, there will be a point to represent every possible input point. Utilizing MVN with infinite dimensions allows us to fit functions with infinite parameters for regression tasks, thus enabling predictions throughout the region of interest. In Fig. \ref{FIG:8}, We illustrate 200 samples from a two hundred-variate normal (200-VN) distribution to conceptualize functions with infinite parameters. We call these functions ``kernelized prior functions'', because there are no observed data points yet. All functions are randomly generated by the MVN model with kernel functions as prior knowledge before having any observed data points. 
\subsection{Non-parametric Model}

This section explains the distinction between parametric and non-parametric models \cite{Murphy2012}. Parametric models assume that the data distribution can be modeled in terms of a set of finite numbers of parameters. In regression, given some data points, we would like to predict the function value $y=f(x)$ for a new specific $x$. If we assume a linear regression model, $y = \theta_1  + \theta_2 x$, we need to identify the parameters $\theta_1$ and $\theta_2$ to define the function. often, a linear model is insufficient, and a polynomial model with more parameters, like $y = \theta_1+\theta_2 x+\theta_3 x^2$ is needed. We use the training dataset $D$ comprising $n$ observed points, $D=[(x_i,y_i)\, \vert \, i=1, \ldots, n]$ to train the model, i.e. establish a mapping $x$ to $y$ through basis functions $f(x)$. After the training process, all information in the dataset is assumed to be encapsulated by the feature parameters $\mathbf{\theta}$, thus predictions are independent of the training dataset $D$. This can be expressed as $P(f_* \, \vert \,  X_*, \mathbf {\theta} ,D)=P(f_* \, \, \vert \,  X_*, \mathbf {\theta})$, in which $f_*$ are predictions made at unobserved data points $X_*$. Thus, when conducting regressions using parametric models, the complexity or flexibility of models is inherently limited by the number of parameters. Conversely, if the parameter number of a model grows with the size of the observed dataset, it’s a non-parametric model. Non-parametric models do not imply that there are no parameters; but rather they entail an infinite number of parameters. 

\section{GAUSSIAN PROCESSES}

Before delving into the Gaussian processes, we first do a quick review of the foundational concepts we have covered. In regression, our objective is to model a function $\mathbf{f}$ based on observed data points $D$ (the training dataset) from the unknown function $\mathbf{f}$. Traditional nonlinear regression methods often give a single function that is considered to best fit the dataset. However, there could be more than one function that fits the observed data points equally well. We observed that when the dimension of MVN was infinite, we could make predictions at any point using these infinite numbers of functions. These functions are MVN because it is our (prior) assumption. More formally, the prior distribution of these infinite functions is MVN, representing the expected outputs of $\mathbf{f}$ over inputs $\mathbf{x}$ before observing any data. When we start to have observations, instead of infinite numbers of functions, we only keep functions that fit the observed data points, forming the posterior distribution. This posterior is the prior updated with observed data. When we have new observations, we use the current posterior as prior, and new observed data points to obtain a fresh posterior.  

\textbf{Definition of Gaussian processes}: A Gaussian process model describes a probability distribution over possible functions that fit a set of points. Because we have the probability distribution over all possible functions, we can compute the means to represent the maximum likelihood estimate of the function, and the variances as an indicator of prediction confidence. Key points include: i) the function prior is updated with new observations; ii) a Gaussian process model is a probability distribution over possible functions, with any finite samples of functions being jointly Gaussian distributed; iii) the mean function derived from the posterior distribution of possible functions \textit{is} the function used for regression predictions.

\begin{figure}
	\centering
		\includegraphics[width=0.5\columnwidth]{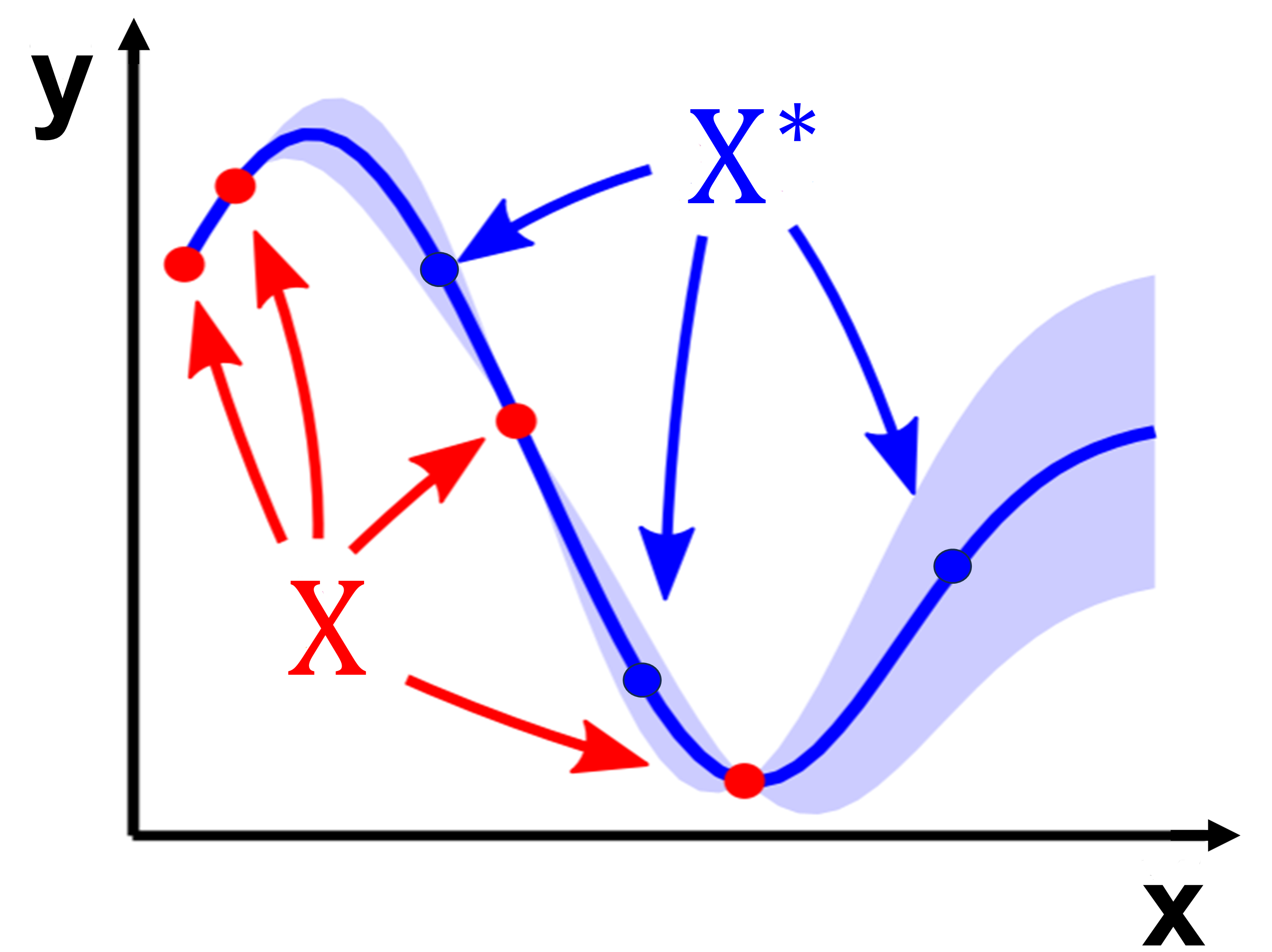}
	\caption{A illustrative process of conducting regressions by Gaussian processes. The red points are observed data, the blue line represents the mean function estimated by the observed data points, and predictions will be made at new blue points.}
	\label{FIG:9}
\end{figure}

Now, it is time to explore the standard Gaussian process model. All the parameter definitions align the classic textbook by Rasmussen (2006) \cite{Rasmussen2006}. Besides the covered basic concepts, Appendix A.1 and A.2 of \cite{Rasmussen2006} are also recommended reading. The regression function modeled by a multivariate Gaussian is given by:
\begin{ceqn}
    \begin{align}
       P(\mathbf{f} \, \lvert\, \mathbf{X}) = \mathcal{N}(\mathbf{f} \, \lvert\, \boldsymbol\mu, \mathbf{K}) \ , \nonumber
    \end{align}
\end{ceqn}
where $\mathbf{X} = [\mathbf{x}_1, \ldots, \mathbf{x}_n ]$ represents the observed data points, $\mathbf{f} = \left[ f(\mathbf{x}_1), \ldots, f(\mathbf{x}_n) \right]$ the function values, $\boldsymbol\mu = \left[ m(\mathbf{x}_1), \ldots, m(\mathbf{x}_n) \right]$ the mean function, and $K_{ij} = k(\mathbf{x}_i,\mathbf{x}_j)$ the kernel function, which is a positive definite. With no observation, we default the mean function to $m(\mathbf{X}) = 0$, assuming the data is normalized to zero mean. The Gaussian process model is thus a distribution over functions whose shapes (smoothness) are defined by $\mathbf{K}$. If points $\mathbf{x}_i$ and $\mathbf{x}_j$ are considered similar by the kernel,  their respective function outputs, $f(\mathbf{x}_i)$ and $f(\mathbf{x}_j)$, are expected to be similar too. The regression process using Gaussian processes is illustrated in Fig. \ref{FIG:9}: given observed data (red points) and a mean function $\mathbf{f}$ (blue line) estimated from these observed data points, we predict at new points $\mathbf{X}_*$ as $\mathbf{f}(\mathbf{X}_*)$.

The joint distribution of $\mathbf{f}$ and $\mathbf{f}_*$ is expressed as:
\begin{ceqn}
    \begin{align}
       \begin{bmatrix}\mathbf{f} \\ \mathbf{f}_*\end{bmatrix} \sim \mathcal{N}\left(\begin{bmatrix}m(\mathbf{X})\\ m(\mathbf{X}_*)\end{bmatrix}, \begin{bmatrix}\mathbf{K} & \mathbf{K}_* \\ \mathbf{K}_*^\mathsf{T} & \mathbf{K}_{**}\end{bmatrix}\right) \ , \nonumber
    \end{align}
\end{ceqn}
where $\mathbf{K}=K(\mathbf{X}, \mathbf{X})$, $\mathbf{K}_* = K(\mathbf{X}, \mathbf{X}_*)$ and $\mathbf{K}_{**}=K(\mathbf{X}_*, \mathbf{X}_*)$. The mean is assumed to be $\begin{pmatrix}m(\mathbf{X}), m(\mathbf{X}_*)\end{pmatrix} = \mathbf{0}$. 

While this equation describes the joint probability distribution $P(\mathbf{f}, \mathbf{f}_* \, \vert \, \mathbf{X}, \mathbf{X}_*)$ over $\mathbf{f}$ and $\mathbf{f}_*$, in regressions, we need the conditional distribution $P(\mathbf{f}_* \, \vert \, \mathbf{f}, \mathbf{X}, \mathbf{X}_*)$ over $\mathbf{f}_*$ only. The derivation of the conditional distribution $P(\mathbf{f}_* \, \vert \, \mathbf{f}, \mathbf{X}, \mathbf{X}_*)$ from the joint distribution $P(\mathbf{f}, \mathbf{f}_* \, \vert \, \mathbf{X}, \mathbf{X}_*)$ is achieved by using the Marginal and conditional distributions of MVN theorem \cite[Sec.\ 2.3.1]{bishop2006pattern}. The result is:
\begin{ceqn}
    \begin{align}
       \mathbf{f}_* \, \vert \, \mathbf{f}, \mathbf{X}, \mathbf{X}_* \sim \mathcal{N} \left(\mathbf{K}_*^\mathsf{T} \, \mathbf{K}^{-1} \, \mathbf{f}, \: \mathbf{K}_{**} - \mathbf{K}_*^\mathsf{T} \, \mathbf{K}^{-1} \, \mathbf{K}_* \right) \ . \nonumber
    \end{align}
\end{ceqn}

In realistic scenarios, we typically have access only to noisy versions of true function values, $y = f(x) + \epsilon$, where $\epsilon$ represents additive independent and identically distributed (i.i.d.) Gaussian noise with variance $\sigma_n^2$. The prior on these noisy observations then becomes $\text{cov}(y) = \mathbf{K} + \sigma_n^2 \mathbf{I}$. The joint distribution of the observed values and the function values at new testing points is:
\begin{ceqn}
    \begin{align}
       \begin{pmatrix}\mathbf{y} \\ \mathbf{f}_*\end{pmatrix} \sim\mathcal{N}\left(\mathbf{0}, \begin{bmatrix}\mathbf{K} + \sigma_n^2 \mathbf{I} & \mathbf{K}_* \\ \mathbf{K}_*^\mathsf{T} & \mathbf{K}_{**}\end{bmatrix}\right) \ . \nonumber
    \end{align}
\end{ceqn}
By deriving the conditional distribution, we get the predictive equations for Gaussian process regression:
\begin{ceqn}
    \begin{align}
       \mathbf{\bar{f}_*} \, \vert \, \mathbf{X}, \mathbf{y}, \mathbf{X}_* \sim \mathcal{N} \left(\mathbf{\bar{f}_*}, \text{cov}(\mathbf{f}_*)\right) \ , \nonumber
    \end{align}
\end{ceqn}
where
\begin{ceqn}
    \begin{align}
        \mathbf{\bar{f}_*} &\overset{\Delta}{=} \mathbb{E} [\mathbf{\bar{f}_*} \, \vert \, \mathbf{X}, \mathbf{y}, \mathbf{X}_*] \nonumber\\ &= \mathbf{K}_*^\mathsf{T} [\mathbf{K} + \sigma_n^2 \mathbf{I}]^{-1} \mathbf{y} \ , \nonumber \\
        \text{cov}(\mathbf{f}_*) &= \mathbf{K}_{**} - \mathbf{K}_*^\mathsf{T} [\mathbf{K} + \sigma_n^2 \mathbf{I}]^{-1} \mathbf{K}_* \ . \nonumber
    \end{align}
\end{ceqn}
In this expression, the variance function $\text{cov}(\mathbf{f}_*)$ reveals that the uncertainty in predictions depends solely on the input values $\mathbf{X}$ and $\mathbf{X}_*$, not on the observed outputs $\mathbf{y}$. This characteristic is a distinctive property of Gaussian distributions \cite{Rasmussen2006}.

\section{ILLUSTRATIVE EXAMPLE}
\label{sec:example}

This section demonstrates an implementation of the standard GPR, adhering to the algorithm outlined in Rasmussen (2006) \cite[Algorithm 2.1]{Rasmussen2006}.
\begin{ceqn}
    \begin{equation}
      \boxed{\begin{split}
        L &= \text{cholesky}(\mathbf{K} + \sigma^2_n \mathbf{I}) \\
        \boldsymbol{\alpha} &= L^{\top} \setminus (L \setminus \mathbf{y}) \\
    	\mathbf{\bar{f}_*} &= \mathbf{K}_{*}^{\top} \boldsymbol{\alpha} \\
    	\mathbf{v} &= L \setminus \mathbf{K}_{*} \\
    	\mathbb{V}[\mathbf{\bar{f}_*}] &= K(\mathbf{X}_*, \mathbf{X}_*) - \mathbf{v}^{\top} \mathbf{v}.\\ 
    	\log p(\mathbf{y} \mid \mathbf{X}) &= -\frac{1}{2} \mathbf{y}^{\top} (\mathbf{K} + \sigma_n^2 \mathbf{I})^{-1} \mathbf{y} - \frac{1}{2} \log \det(\mathbf{K} + \sigma_n^2 \mathbf{I}) \\
        &\qquad \qquad\qquad\qquad\quad \, \,  \, \, \! - \frac{n}{2} \log 2 \pi \nonumber 
      \end{split}}
    \end{equation}
\end{ceqn}
The inputs of this algorithm are $\mathbf{X}$ (inputs), $\mathbf{y}$(targets), $K$ (covariance function), $\sigma^2_n$(noise level), and $\mathbf{X_{*}}$(test input). The outputs include $\mathbf{\bar{f}_*}$ (mean), $\mathbb{V}[\mathbf{\bar{f}_*}]$ (variance), and $\log p(\mathbf{y} \mid \mathbf{X})$ (log marginal likelihood). 

\begin{figure}
	\centering
		{{\includegraphics[trim=4.5cm 1.5cm 4.4cm 1.0cm, width=\columnwidth]{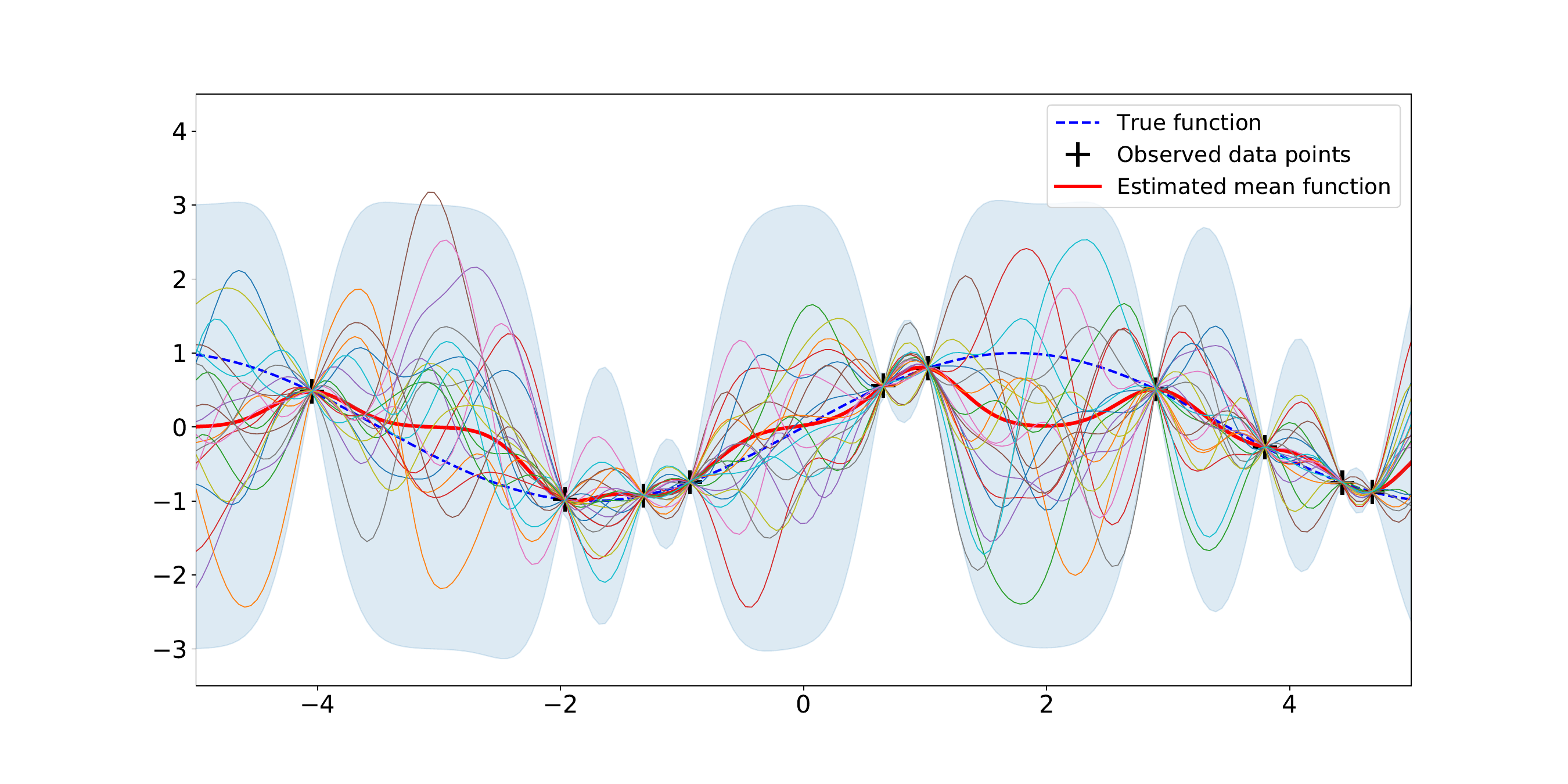}}}
	\caption{An illustrative example of standard GPR. Black crosses represent observed data points generated by the blue dotted line (true function). Given these data points, infinite possible posterior functions were obtained, with 20 samples plotted in different colors. The mean function, derived from the probability distribution of these functions, is plotted as a red solid line. The blue shaded area around the mean function indicates 3 times prediction variances.}
	\label{FIG:10}
\end{figure}

An example result is illustrated in Fig. \ref{FIG:10}. We conducted regression within the [-5, 5] interval. Observed data points (training dataset) were generated from a uniform distribution between -5 and 5. The functions were evaluated at evenly spaced points between -5 and 5. The regression function is composed of mean values estimated by a GPR model. Twenty samples of posterior mean functions, along with 3 times variances, were also plotted. 

\subsection{Hyperparameters Optimization}
\label{sec:Hyperparameters}

\begin{figure}
    \centering
    \subfloat[$l$ = small]{{\includegraphics[trim=3.0cm 1.0cm 3.8cm 2cm, width=0.65\columnwidth]{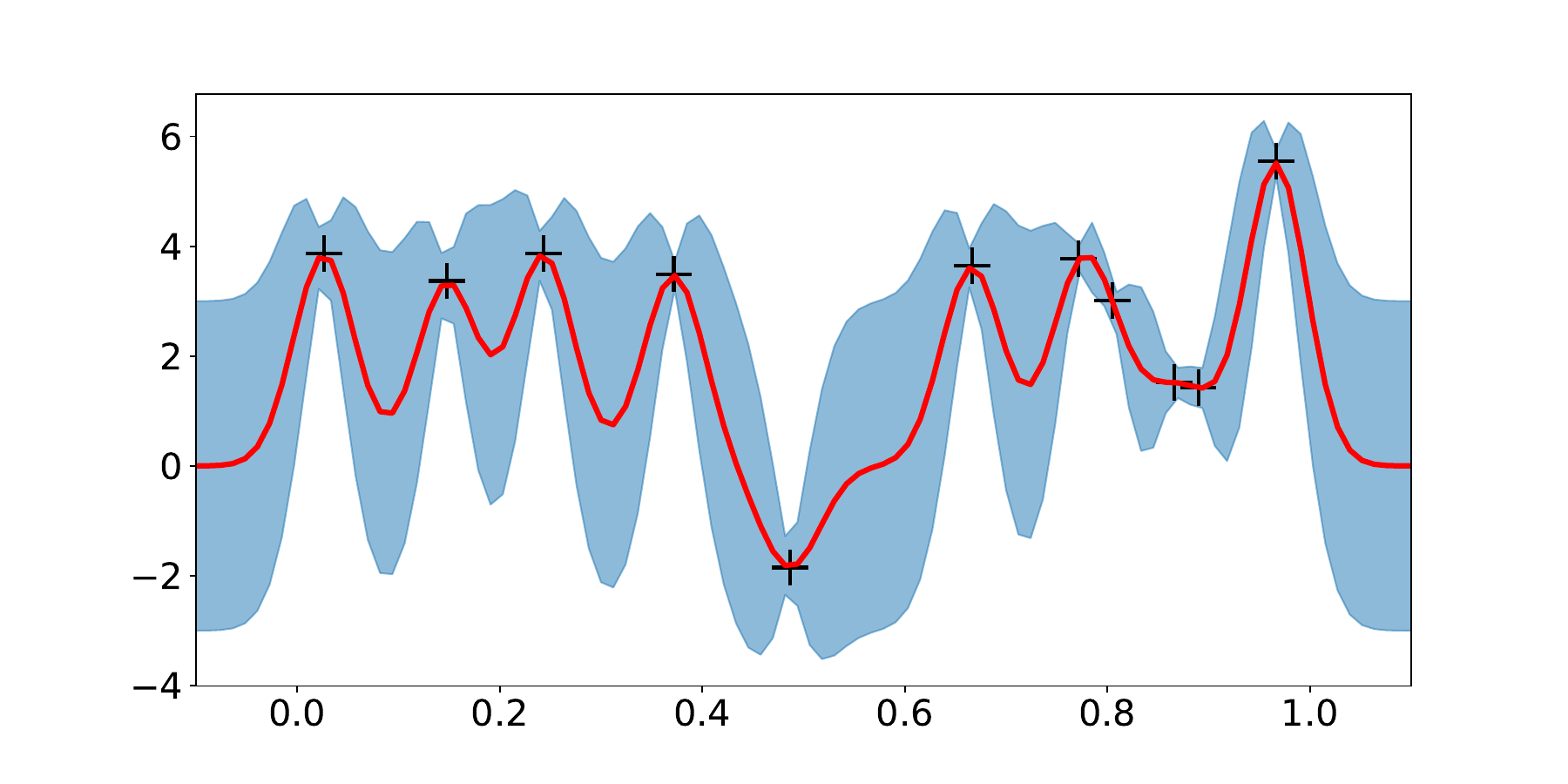} }}
    \qquad
    \vspace{0.1cm}
    \subfloat[$l$ = medium]{{\includegraphics[trim=2.4cm 1.0cm 3.2cm 1.2cm, width=0.68\columnwidth]{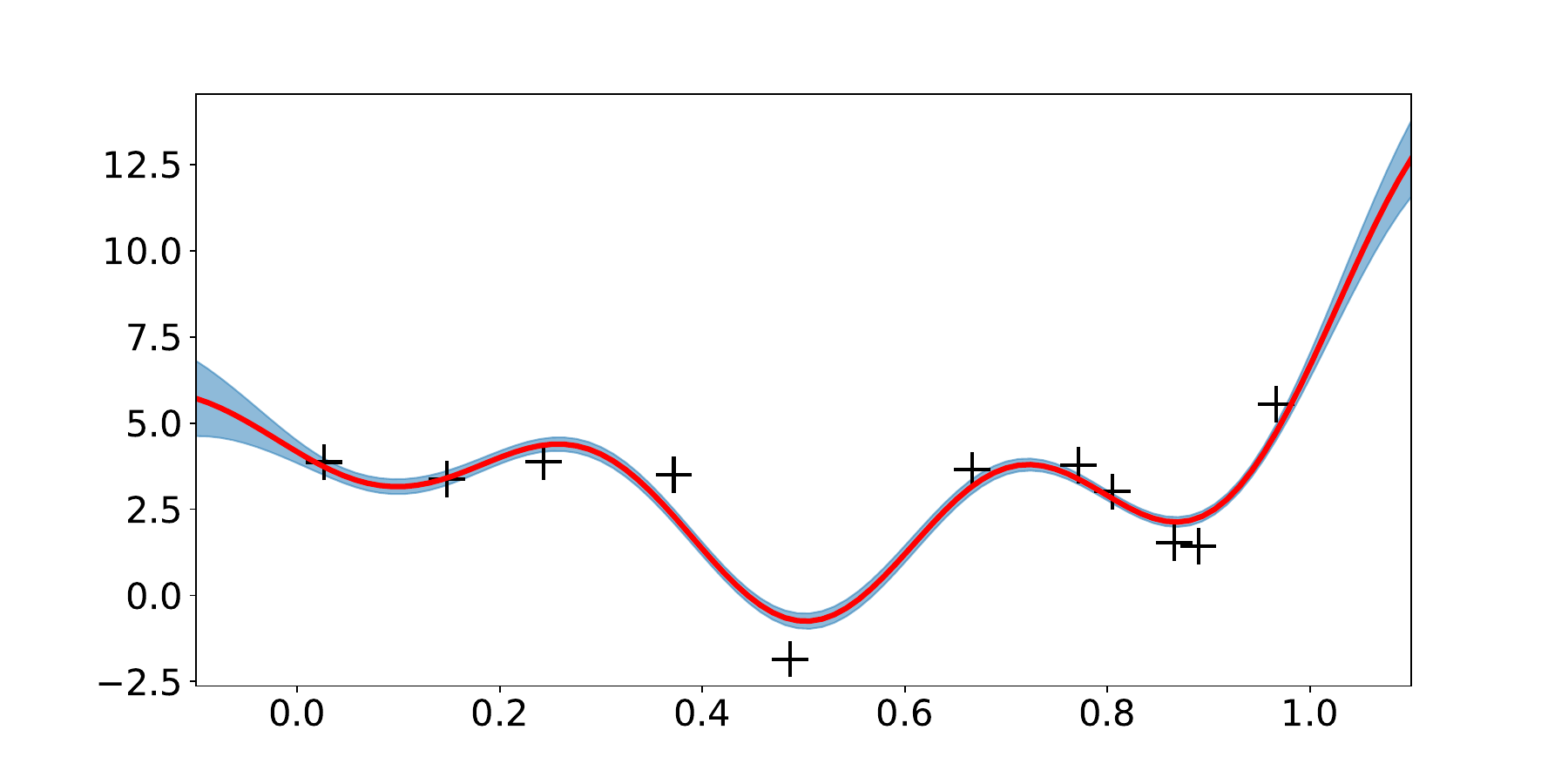} }}
    \qquad 
    \\
    \vspace{0.1cm}
    \subfloat[$l$ = large]{{\includegraphics[trim=3.0cm 1.0cm 3.8cm 2cm, width=0.65\columnwidth]{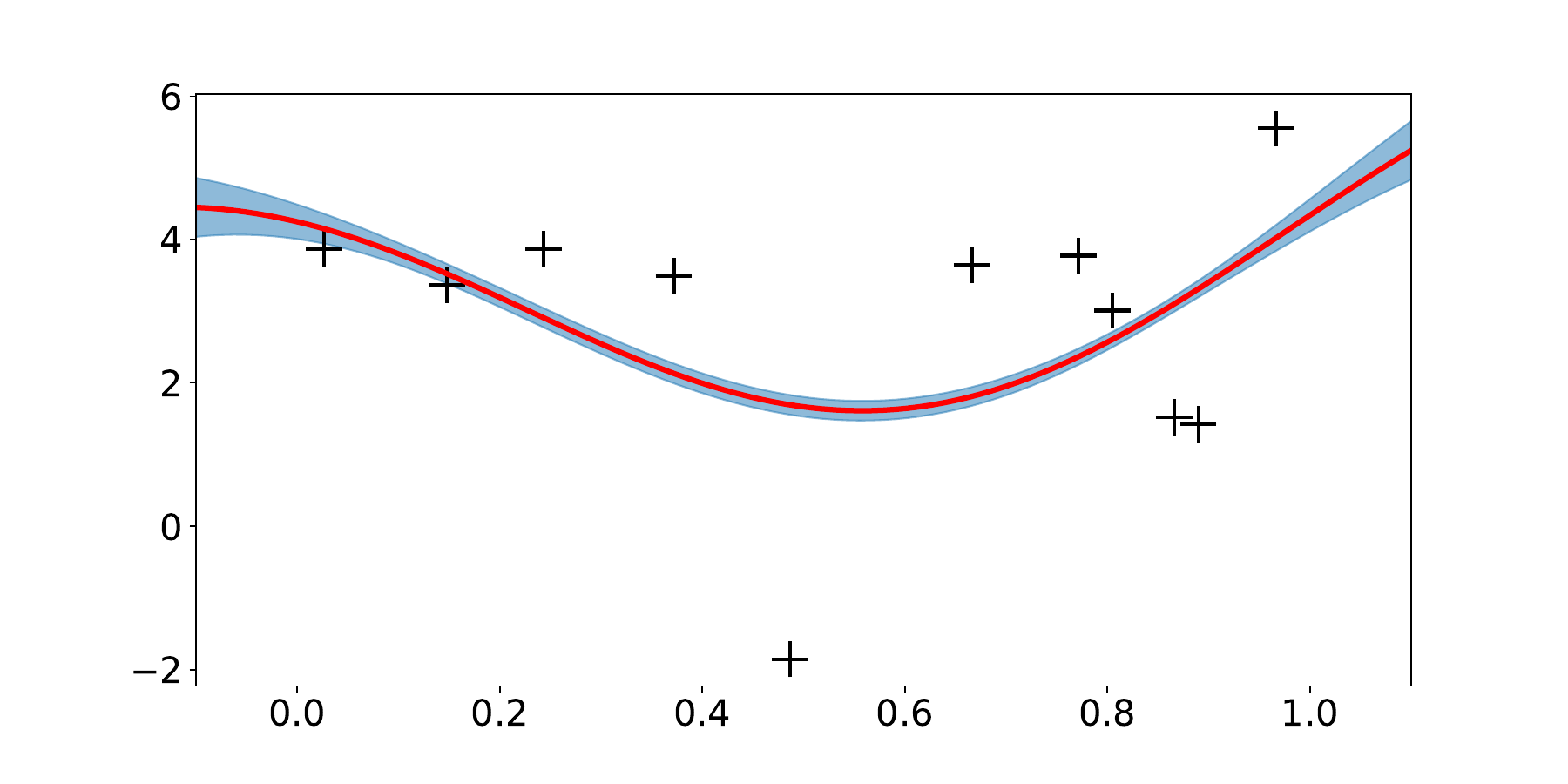} }}
    \caption{Effect of the hyperparameter $l$ on function smoothness: A larger $l$ yields a smoother function, while a smaller $l$ produces a more wiggly function.}
    \label{FIG:11}
\end{figure}

\begin{figure}
	\centering
		{{\includegraphics[trim=2.2cm 1.0cm 2.8cm 0.8cm, width=0.95\columnwidth]{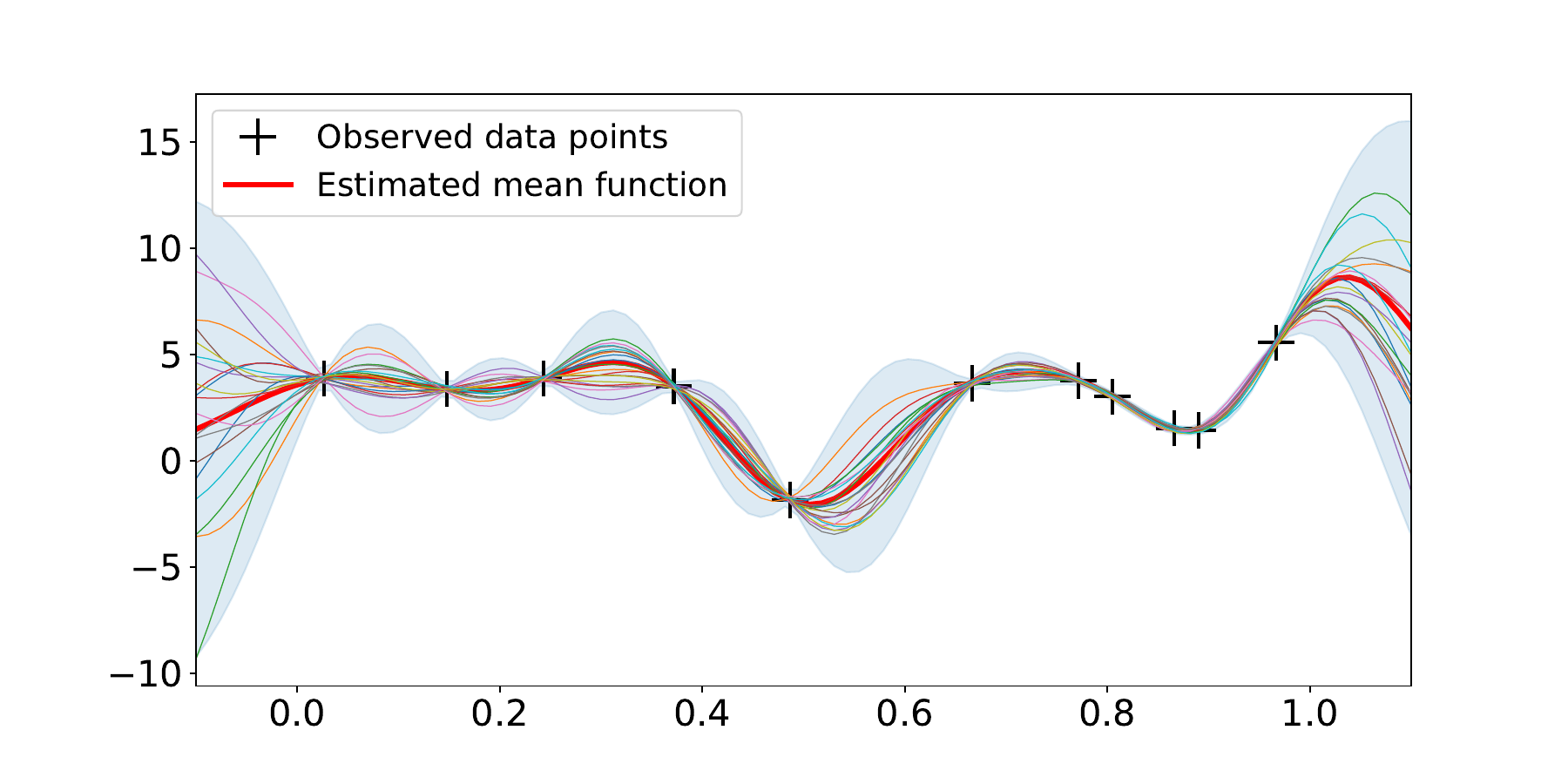}}}
	\caption{Regression result with the optimized hyperparameters $\sigma_f$ and $l$.}
	\label{FIG:12}
\end{figure}

We have covered the basics of GPR and provided a straightforward example. However, practical GPR models often present more complexity. The selection of the kernel function is critical, as it greatly affects the model's ability to generalize \cite{duvenaud2016kernel}. Kernel functions range from well-established options like the RBF to custom designs tailored to specific needs based on model requirements such as smoothness, sparsity, drastic changes, and differentiability \cite{duvenaud2014automatic}. Selecting an appropriate kernel function for a specific GPR task is detailed in Duvenaud (2014) \cite{duvenaud2014automatic}. Additionally, hyperparameter optimization plays an essential role in kernel-based methods. For example, consider the widely used RBF kernel:

\begin{ceqn}
    \begin{align}
        k(\mathbf{x}_i,\mathbf{x}_j) = \sigma_f^2 \exp \Big(-\frac{1}{2l}
         (\mathbf{x}_i - \mathbf{x}_j)^\mathsf{T}
         (\mathbf{x}_i - \mathbf{x}_j) \Big)  \ , \nonumber
    \end{align}
\end{ceqn}
In this kernel, $\sigma_f$ (vertical scale) and $l$ (horizontal scale) are hyperparameters. The parameter $\sigma_f$ determines the vertical span of the function, while $l$ indicates the rate at which the correlation between two points decreases with increasing distance. The influence of the hyperparameter $l$ on the smoothness of the function is demonstrated in Fig. \ref{FIG:11}. Increasing the value of $l$ results in a smoother function, while a smaller $l$ value leads to a function with more fluctuations or `wiggles'.
The optimal hyperparameters $\boldsymbol{\Theta^*}$ are determined by maximizing the log marginal likelihood \cite{Rasmussen2006}:
\begin{ceqn}
    \begin{align}
        \boldsymbol{\Theta^*} = \arg\max\limits_{\Theta} \log p(\mathbf{y} \, \vert \, \mathbf{X}, \boldsymbol{\Theta})  \ . \nonumber
    \end{align}
\end{ceqn}
Thus, considering hyperparameters, a more generalized prediction equation at new testing points is \cite{gpss2019}:
\begin{ceqn}
    \begin{align}
        \mathbf{\bar{f}_*} \, \vert \, \mathbf{X}, \mathbf{y}, \mathbf{X}_*,  \boldsymbol{\Theta} \sim \mathcal{N} \left(\mathbf{\bar{f}_*}, \text{cov}(\mathbf{f}_*)\right)  \ . \nonumber
    \end{align}
\end{ceqn}
Note that after learning/optimizing the hyperparameters, the predictive variance $\text{cov}(\mathbf{f}_*)$ depends on not only the inputs $\mathbf{X}$ and $\mathbf{X}_*$ but also the outputs $\mathbf{y}$ \cite{chen2018priors}. With the optimized hyperparameters, $\sigma_f = 0.0067$ and $l = 0.0967$, the regression result of the observed data points shown in Fig. \ref{FIG:11} is depicted in Fig. \ref{FIG:12}. Here, the hyperparameters optimization was conducted by the GPy package, which will be introduced in the next section. 
\subsection{Gaussian Processes Packages}
This section reviews three Python packages for implementing Gaussian processes. GPy is a mature and well-documented package in development since 2012 \cite{de2017gpflow}. It utilizes NumPy for computations, offering sufficient stability for tasks that are not computationally intensive. However, GPR is computationally expensive in high dimensional spaces (beyond a few dozen). For complex and computationally intense tasks, packages incorporating advanced algorithms and GPU acceleration are especially preferable. GPflow \cite{de2017gpflow} originates from GPy with a similar interface. It leverages TensorFlow as its computational backend. GPyTorch \cite{gardner2018gpytorch} is a more recent package that provides GPU acceleration through PyTorch. Like GPflow, GPyTorch supports automatic gradients, which simplifies the development of complex models, such as those embedding deep neural networks within GP frameworks.

\section{CONCLUSION}
A Gaussian process is a probability distribution over possible functions that fit a set of points \cite{Rasmussen2006}. A Gaussian process regression model provides prediction values together with uncertainty estimates. The model incorporates prior knowledge about the nature of the functions through the use of kernel functions.

The GPR model discussed in this tutorial is the standard or ``vanilla'' approach to Gaussian processes \cite{frigola2013bayesian}. There are two primary limitations with it: 1) The computational complexity is $O(N^3)$, where $N$ represents the dimension of the covariance matrix $K$. 2) The memory consumption increases quadratically with data size. Due to these constraints, standard GPR models become impractical for large datasets. In such cases, sparse Gaussian Processes are employed to alleviate computational complexity \cite{liu2020gaussian}.

\section{ACKNOWLEDGMENTS}
The author would like to express sincere gratitude to Prof. Krzysztof Czarnecki from the University of Waterloo. His insightful and constructive feedback have significantly contributed to the progression and enhancement of the quality of this tutorial. The author is deeply thankful for his invaluable guidance and support.

\def\refname{REFERENCES}

\bibliographystyle{IEEEtran}
\bibliography{bibliography}

\begin{IEEEbiography}{Jie Wang}{\,}is a postdoctoral research associate at the University of Waterloo. He earned his Ph.D. degree in Mechanical Engineering from the University of Calgary. His research bridges machine learning and traditional robotics, primarily focusing on enhancing the performance of mobile robots in real-world settings while ensuring safety and efficiency. For further information or collaboration inquiries, he can be reached at jwangjie@outlook.com. \vadjust{\vfill\pagebreak}
\end{IEEEbiography}

\end{document}